\documentclass[fleqn,10pt]{wlscirep}
\usepackage[utf8]{inputenc}
\usepackage{verbatim}
\usepackage[T1]{fontenc}
\usepackage{multirow}
\usepackage{diagbox}
\title{MPSA-DenseNet: A novel deep learning model for English accent classification}

\author[1]{Tianyu Song}
\author[2]{Linh Thi Hoai Nguyen}
\author[1,*]{Ton Viet Ta}
\affil[1]{Mathematical Modeling Laboratory, Kyushu University, Fukuoka 819-0395, Japan}
\affil[2]{Institute of Mathematics for Industry, Kyushu University, Fukuoka 819-0395, Japan}

\begin{abstract}

This paper presents three innovative deep learning models for English accent classification: Multi-DenseNet, PSA-DenseNet, and MPSE-DenseNet, that combine multi-task learning and the PSA module attention mechanism with DenseNet. We applied these models to data collected from six dialects of English across native English speaking regions (Britain, the United States, Scotland) and nonnative English speaking regions (China, Germany, India). Our experimental results show a significant improvement in classification accuracy, particularly with MPSA-DenseNet, which outperforms all other models, including DenseNet and EPSA models previously used for accent identification. Our findings indicate that MPSA-DenseNet is a highly promising model for accurately identifying English accents.
\end{abstract}
\begin{document}

\flushbottom
\maketitle
%
%
\thispagestyle{empty}

\section{Introduction}

In recent years, deep learning technology has made significant advancements in the field of speech applications, among which accent recognition has received much attention. Accents are variate in the emphasis on words or syllables, and are often influenced by a person's upbringing or social background. 

As the most widely used language in the world, English plays a crucial role in international communication and business activities. However, different usage environments, historical and cultural backgrounds in different regions, have resulted in various accents and dialects of English, increasing the difficulty of recognizing correct content by smart voice devices. Therefore, there is a need to enhance the recognition of different accents. 

In the field of accent classification, feature extraction of speech signals plays a critical role. Mel-scale Frequency Cepstral Coefficients (MFCC) is a widely used method for speech signal feature extraction, which has been confirmed to have good performance in accent classification by many studies \cite{ICY,HMG,AAR}. Traditional machine learning methods such as Support Vector Machine (SVM) \cite{PD,RA,TG,HLZ}, Hidden Markov Model (HMMs) \cite{AH,KK} were used before the emergence of deep learning. Studies by Carol Pedersen and Joachim Diederich \cite{PD}, Hong Tang and Ali A. Ghorba \cite{TG}, and Jordan J. Bird et al \cite{BWE}. showed good classification results using SVM, Pairwise SVM, and $K$-Nearest Neighbor (KNN) methods, respectively.

With the development of deep learning, neural network models have gradually replaced traditional machine learning methods in accent recognition and classification research. Deep learning models can automatically extract features from large-scale speech data, thereby improve accent recognition ability. 

Studies by Yishan Jiao et al \cite{YMV}. and Keven Chionh et al \cite{KMY}. showed that combining Deep Neural Network (DNN) and Recurrent Neural Network (RNN) or using Convolutional Neural Network (CNN) greatly improved the classification accuracy of accents. Additionally, research by Al-Jumaili et al \cite{ZTA}. showed that using transfer learning with a lightweight neural network model achieved high-precision results on accent classification tasks.

Furthermore, attention mechanism \cite{QHY} and multi-task \cite{YHD} learning have also been applied in the speech and audio fields in recent years, showing improved classification performance. These methods have been verified to effectively improve the classification performance of the original basic model in experiments.

While most of the current research focuses on accurately translating speech signals with different accents into correct content, our study focuses on the recognition and classification of different accents based on the country or region of the speaker's English accent. In this study, we newly introduce three novel deep learning models named Multi-DenseNet, PSA-DenseNet, and MPSA-DenseNet, which combine DenseNet \cite{GZL}, multi-task, and attention mechanism models, to achieve better performance compared to previous models such as DenseNet or ResNet for accent identification.
This study has an important impact on communication, language learning, and social interactions. It can also have practical applications in areas such as speech recognition, language translation, and forensic linguistics.

The remaining of this paper is organized as follows. The Materials and Methods section describes the architectures of the three new deep learning models: Multi-DenseNet, PSA-DenseNet, and MPSA-DenseNet. In the Results section, we validate the effectiveness of our proposed models by comparing them with existing models. Finally, the Conclusion section provides concluding remarks and outlines future research directions.
 
\section{Methods and Materials}
In this section, first we describe data collection and preprocessing process, which involve processing raw speech data and converting it into spectrograms. Second, we introduce our deep learning models, Multi-DenseNet, PSA-DenseNet, and MPSA-DenseNet, which combines DenseNet, multi-task, and attention mechanism models. Evaluation metrics and GPU configuration used in the study are also given.

\subsection{Data collection}
In this research, the dataset is composed of partial data screened from the Speech Accent Archive \cite{SAA}, the CSTR VCTK Corpus \cite{VCTK}, and some self-collected speech datasets.

English accent classification experiments have been conducted using various datasets, focusing primarily on non-native-speaking countries and regions with some studies examining geographically close accents. To ensure dataset diversity, we avoided selecting a single type of speech data as much as possible, resulting in six dialects of English in the dataset, including three native English speaking regions (Britain, the United States, and Scottish) and three nonnative English speaking regions (China, Germany, and India). 

In addition, since we utilized a multi-task learning method, age and gender categories as auxiliary tasks are included. This approach aimed to train a model with better robustness. The age category is divided into five groups: below 20, 20-29, 30-39, 40-49, and 50 and above, which are respectively numbered from 0 to 4. The gender category is grouped where male is numbered 0, and female is numbered 1.

\begin{figure}[ht]
	\begin{center}
			\includegraphics[width=0.33\textwidth]{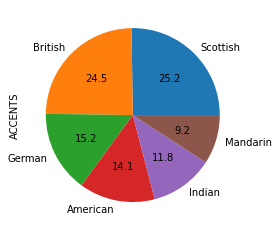}
			\includegraphics[width=0.3\textwidth]{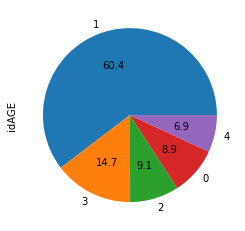}
			\includegraphics[width=0.3\textwidth]{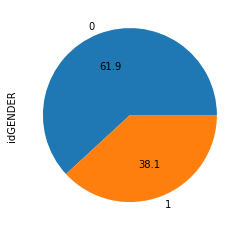}
	\caption{%
		Left: Percentage of data for accent categories; Middle: Percentage of five age groups: below 20, 20-29, 30-39, 40-49, and 50 and above, which are numbered from 0 to 4, respectively; Right: Percentage of male (numbered 0) and female (numbered 1).
	}
	\label{figure1}
	\end{center}
\end{figure}

The dataset consists of 9117 samples with the proportions for dialects, age groups and genders shown respectively in three Pie graphs in Figures~\ref{figure1}. 

\subsection{Data preprocessing}
In this section, we describe step by step the speed data preprocessing flow.

\subsubsection{File format unifying}
Different formats of the audio files, such as wav, mp3, and mp4 are converted into a unified wav format that can be processed using Librosa \cite{MRL} --an audio processing Python library. The FFmpeg \cite{TS} framework has been used to prevent any unnecessary impacts on the voice files during conversion and to achieve lossless conversion to the greatest extend.

\subsubsection{Standardization}
To standardize the voice files, we adjusted the sampling rate to $44 100$, a value frequently used in the field. We converted monophonic audio data into two channels and standardized the duration to an average duration of 6 seconds for all files. To achieve this, we trimmed longer audio files and extended shorter audio files by adding silence.

\subsubsection{Enhancement}
To improve the robustness and generalization ability of the model, we used data enhancement processing. Specifically, we added Gaussian white noise, which is a type of noise generated from normally distributed random values with mean zero and standard deviation $\sigma=10$ to the input data. This process helps the model learn features that are robust to slight changes in the input, therefore decreases the generalization error, that is, the classification error on unseen data.

\subsubsection{Spectrogram conversion}

\begin{figure}[ht]
\centering
\includegraphics[width=\linewidth]{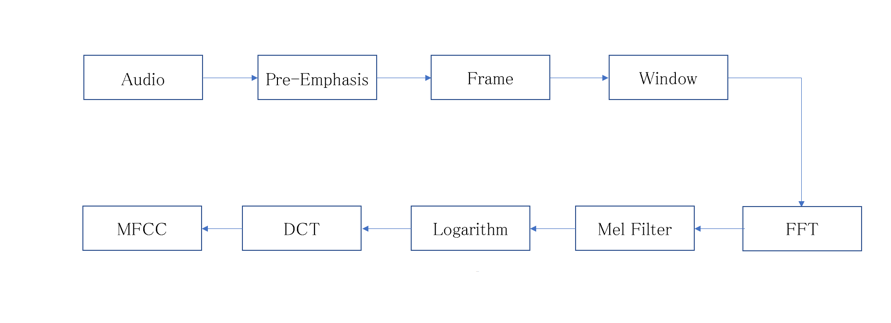}
\caption{Spectrogram conversion process using MFCC  method.}
\label{figure3}
\end{figure}

This subsection explains the MFCC method \cite{ICY}. It is a method in the Librosa 
to convert speech data files into spectrograms. The conversion process is shown in Figure \ref{figure3}.


To enhance the high-frequency part of the speech signal and keep it in the entire frequency band from low frequency to high frequency, each data file is passed through a high-pass filter: 
$$y(n)=x(n)- \alpha \times x(n-1).$$
Here, $y(n)$ and $x(n)$ are the output signal and input signal respectively, $\alpha$ is the pre-emphasis factor. We  take $\alpha=0.97$. 
This ensures that the same signal-to-noise ratio can be used for the frequency spectrum. Each data file is then divided into frames according to a certain fixed time length to obtain short-term stable input. Each small sample after segmentation is called a frame. It is also a sample for subsequent analysis and extraction of MFCC.

To avoid the discontinuity problem caused by framing and improve the continuity at both ends of the sample, window processing is added to each frame after framing. After windowing, the speech signal is converted into an energy distribution in the frequency domain to make it easier to observe the characteristics of the signal. Therefore, fast Fourier transform (FFT) \cite{BE} is performed on each frame signal after framing and windowing to obtain the spectrum of each frame. Then, the energy spectrum passes through the Mel filter to convert the ordinary frequency $f$ of each audio to Mel frequencies closer to our ears. The conversion formula is as follows: 
$$mel(f)= 2995 \times \log_{10} \left(1+\frac{f}{700} \right).$$

\begin{figure}[ht]
\centering
\includegraphics [width=6cm]{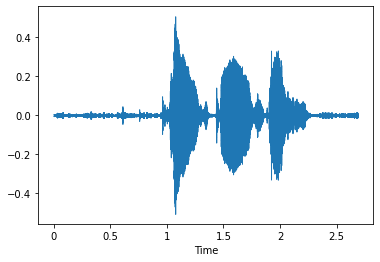} \quad
\includegraphics[width=6 cm]{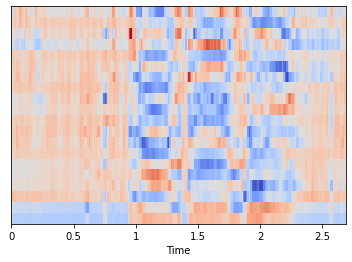} \\
\includegraphics [width=6cm]{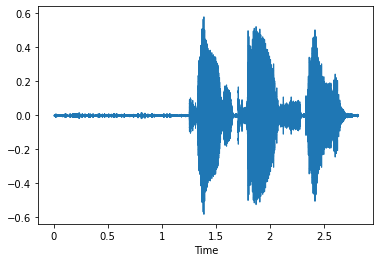}
\includegraphics[width=6 cm]{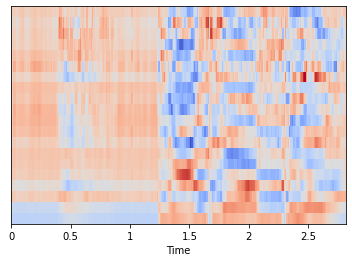}
\caption{Left: the original signals: of British-accent speaker on first row and of English Scottish-accent speaker on the second row; Right: the corresponding MFCC feature maps. }
\label{figure4}
\end{figure}

In Figures \ref{figure4}, the left sub-figures display the original signals of two speech files from British and Scottish speakers, respectively, with the same speech content. The right sub-figures show the corresponding MFCC feature maps of the two signals, demonstrating the effective capture of accent differences between the two speakers by MFCC.

\begin{figure}[ht]
\centering
\includegraphics [width=6cm]{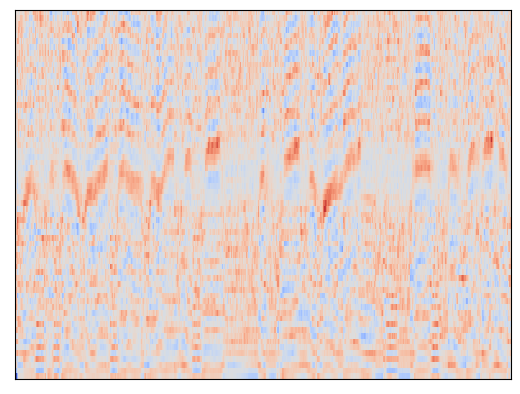}  
\includegraphics [width=6cm]{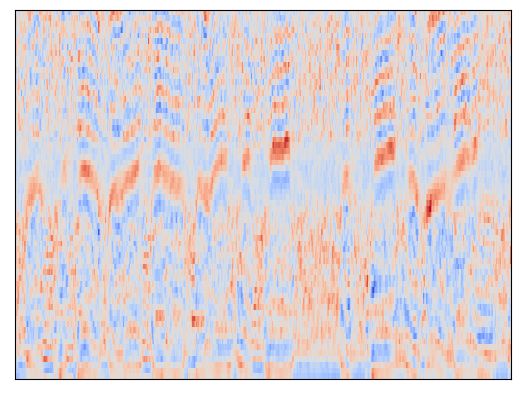}  
\caption{MFCC features without (left) and with (right) speech data pre-processing.}
\label{fig11}
\end{figure}

In addition, Figure \ref{fig11} displays the outputs of the MFCC method with and without speech data preprocessing. The right sub-figure obtained through preprocessing shows a shape of (2, 64, 516), indicating two channels, 64 MFCCs, and 516 frames. The number of frames, $nf$, depends on the length of the speech file and is calculated using the formula 
$$nf = \frac{\text{time} \times \text{sample rate}}{\text{hop length}},$$
with a sample rate of $44100$, a time duration of 6 seconds, and a hop length (i.e. a number of samples between consecutive frames) of 512. Meanwhile, the left sub-figure shows the output of the MFCC method without pre-processing, having a shape of (2, 64, 690). The comparison reveals that the pre-processed MFCC features exhibit clearer and more distinct characteristics compared to the unprocessed features.

\subsection{Deep learning models}

In this subsection, we present our deep learning models: Multi-DenseNet, PSA-DenseNet, and MPSA-DenseNet. The backbone of these three models is DenseNet, which is introduced in detail in the next sub-subsection. 

\subsubsection{DenseNet model}
\begin{figure}[ht]
\centering
\includegraphics[width=14cm]{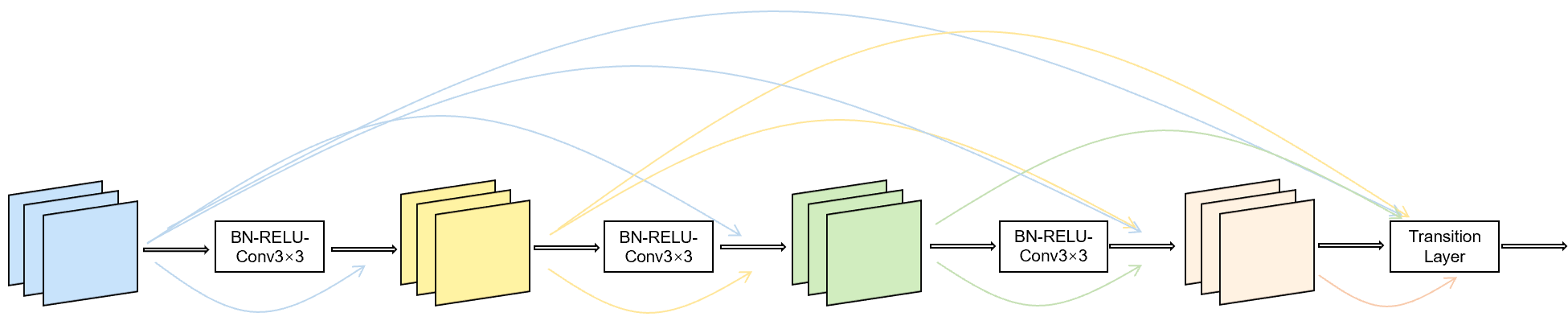}
\caption{Dense blocks and transition layers of DenseNet.}
\label{fig6}
\end{figure}

DenseNet \cite{GZL} comprises two key components: dense blocks and transition layers. The dense blocks determine how the inputs and outputs are concatenated, while the transition layers control the number of channels to prevent it from becoming too large. The structure of the main components is illustrated in Figure \ref{fig6}.

\begin{figure}[ht]
\centering
\includegraphics[width=14cm]{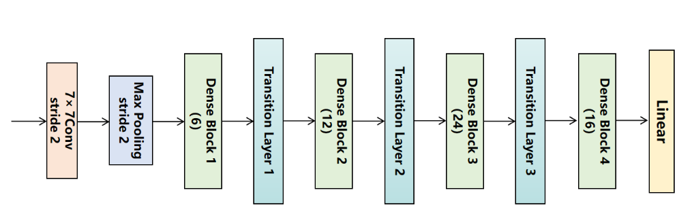}
\caption{DenseNet-121 architecture.}
\label{fig7}
\end{figure}

In DenseNet, each layer is connected to all the preceding layers in a feed-forward manner. For a $K$-layer DenseNet, there are $\frac{K(K+1)}{2}$ connections, and the input to each layer comes from the output of all previous layers. DenseNet has several main architectures, and we selected the DenseNet-121 architecture in this study, as shown in Figure \ref{fig7}.

\subsubsection{Multi-DenseNet model}
\begin{figure}[ht]
\centering
\includegraphics [width=14cm]{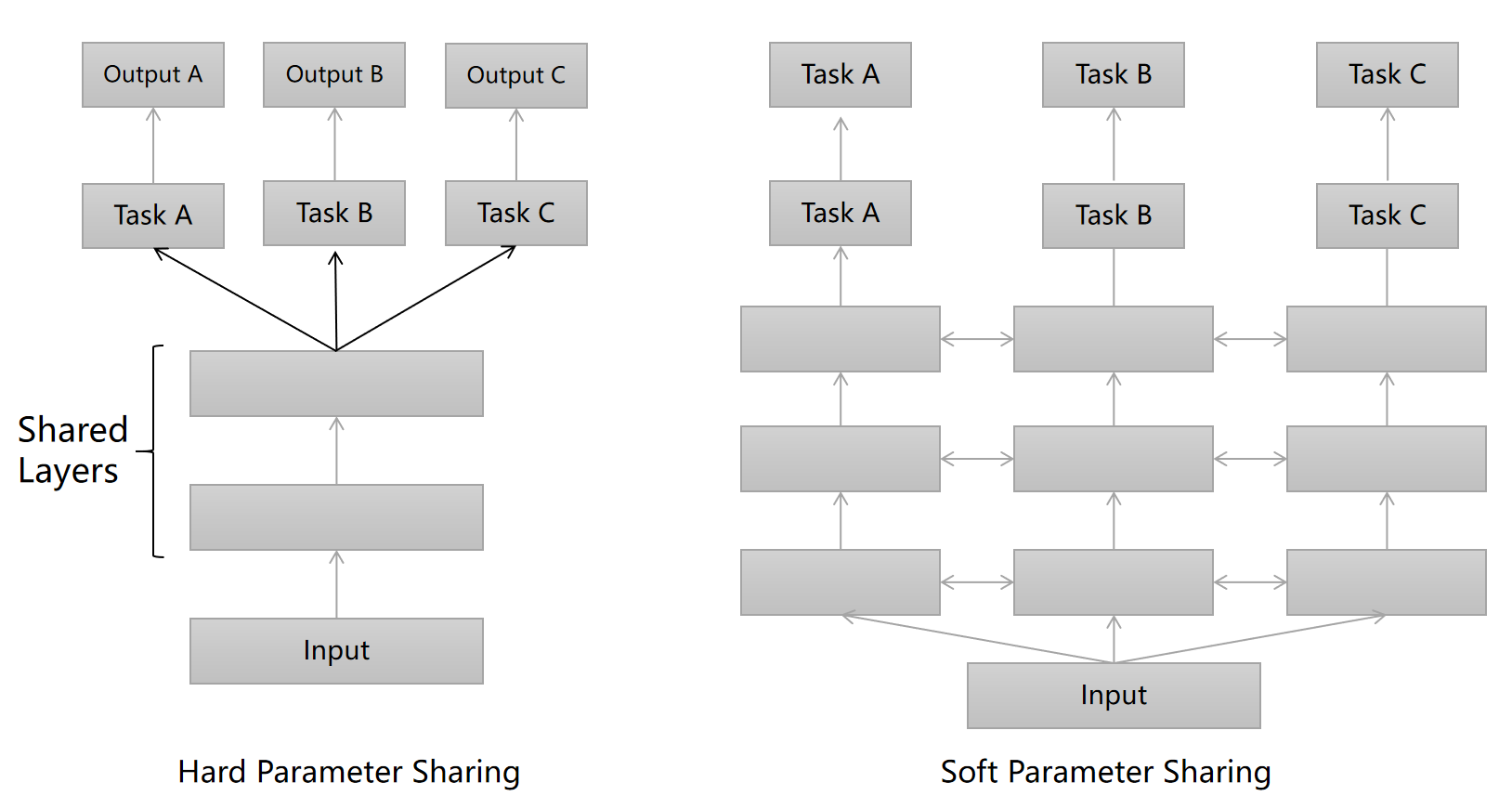}  
\caption{Two main categories of multi-task learning: hard parameter sharing and soft parameter sharing.}
\label{fig8_1}
\end{figure}

The concept of multi-task learning \cite{CR} involves training a neural network to simultaneously learn multiple related tasks. This is accomplished by backpropagating the gradients from all tasks together. By sharing parameters at the bottom layer, the tasks can benefit from each other's learning, thereby enhancing the model's generalization capability. Two main categories of multi-task learning are commonly employed: hard parameter sharing and soft parameter sharing (see Figure \ref{fig8_1}). The distinction lies in how the parameters in the bottom layer are shared among the tasks.

In hard parameter sharing, only a few specific output layers, such as Softmax, are added for each task. The underlying parameters are shared uniformly across all tasks, utilizing a shared backbone. On the other hand, soft parameter sharing entails each task having its own unique model and parameters, while still sharing the underlying parameters.

In our research, we adopt the hard parameter sharing approach as the multi-task learning method. We incorporate age and gender as auxiliary tasks to enhance the accent classification performance. Age and gender exhibit distinct characteristics in speech, such as pronunciation clarity and intonation, which can influence the results of the accent classification task. Additionally, by building a classifier that can determine the age and gender of the speaker's accent, we lay the groundwork for future applications in speaker recognition.

\begin{figure}[ht]
\centering
\includegraphics [width=\linewidth]{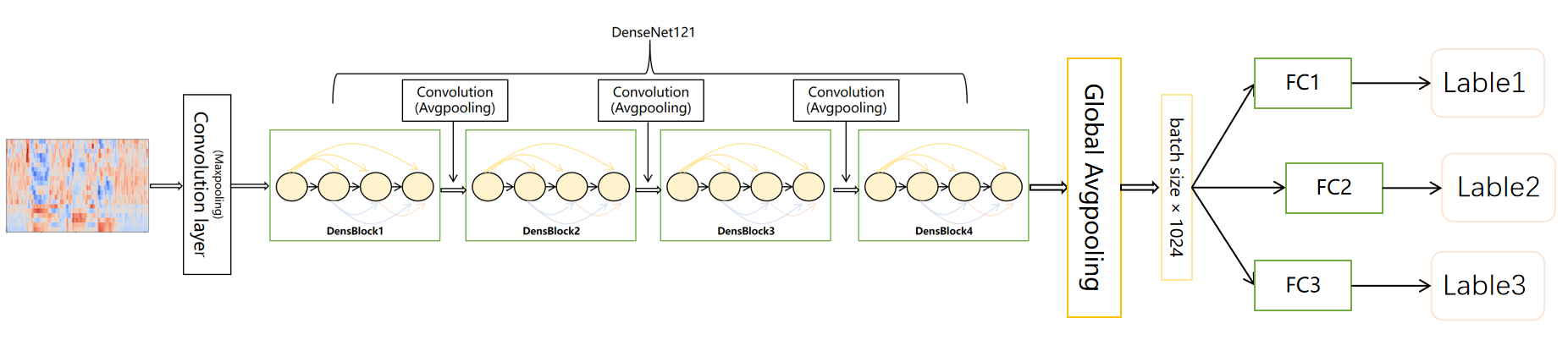}  
\caption{Structure of Multi-DenseNet model.}
\label{fig8_2}
\end{figure}

Our proposed model, Multi-DenseNet, combines DenseNet with multi-task learning. The architecture of Multi-DenseNet is depicted in Figure \ref{fig8_2}.

Next, let's define the total loss function for Multi-DenseNet. Since each task has its own loss function, and the importance of each task may vary, it is necessary to assign appropriate weights to the losses for aggregating them into a total loss function. The simplest approach for assigning multi-task loss weights is to linearly sum the individual task losses. The formula for this method is expressed as:
$$L_{total}=\sum_i \omega_i L_i,$$
 where $L_i$ represents the loss function of each task, and $\omega_i$ denotes the weight assigned to each task. These weights are typically manually set as prior hyperparameters, ensuring that $\sum_i \omega_i=1$.
 
In our study, the cross-entropy \cite{CM} function is adopted as the loss function. Thus, the total loss function is calculated as follows: 

$$L_{total} = \omega_1\sum_{i=1}^{C_1} y_{i}^{(1)} \log(\hat{y}_{i}^{1}) + \omega_2\sum_{i=1}^{C_2} y_{i}^{(2)} \log(\hat{y}_{i}^{2}) + \omega_3\sum_{i=1}^{C_3} y_{i}^{(3)} \log(\hat{y}_{i}^{3}).$$
Here, $y_{i}^{(k)}, \hat{y}_{i}^{k},$ and $C_i$ represent the true label, the predicted value, and the total number of classes for each task, respectively.

\subsubsection{PSA-DenseNet model}
To enhance the selection of important and relevant features, an attention mechanism can be incorporated into a neural network. The integration of an attention mechanism with a CNN has proven to be effective in improving model performance in various studies.

In our study, we utilizes a specific attention mechanism called PSAModule (Pyramid Split Attention Module), derived from EPSANet \cite{HKJ}. The PSAModule consists of two key components: SPCModule and SEWeight module. 

\begin{figure}[ht]
\centering
\includegraphics [width=12cm]{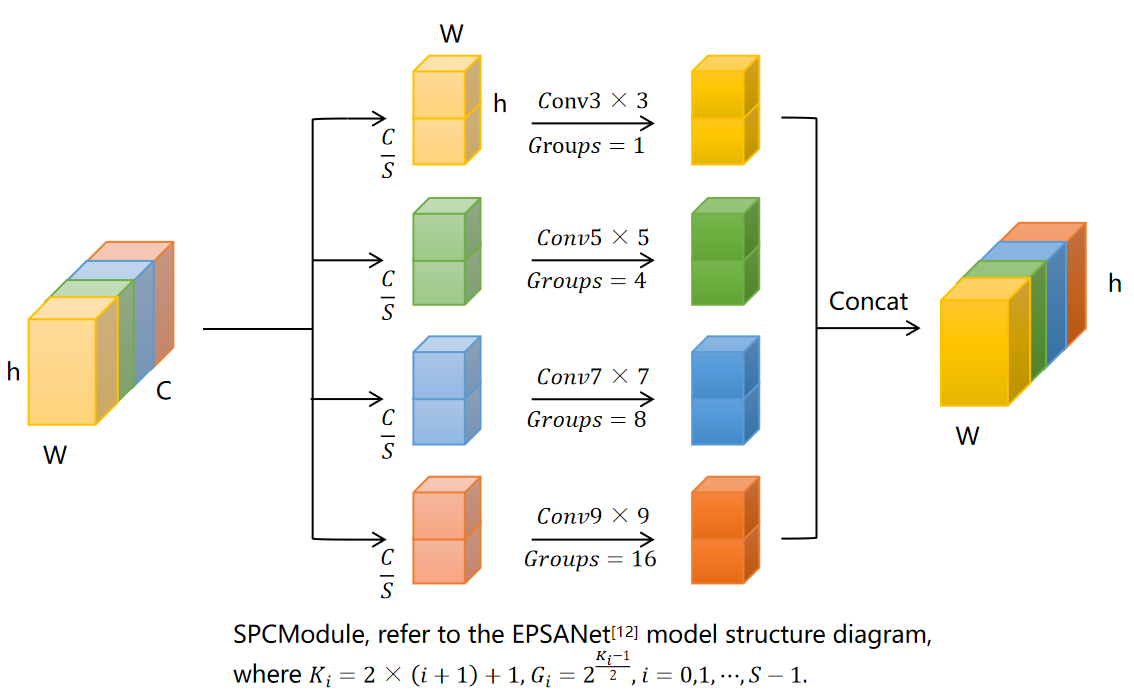}  
\caption{Structure of SPC Module.}
\label{fig8}
\end{figure}

The SPCModule is responsible for channel segmentation and extracting spatial information from each channel to generate multi-scale features. It achieves this by dividing the MFCC feature map into $S$ parts based on the number of input channels ($C$). The resulting feature maps retain the same shape as the original, with $S$ feature maps, each containing $\frac{C}{S}$ channels. The spatial information is then extracted through convolution operations on the feature maps of different scales. By concatenating these multi-scale feature maps, a new multi-scale fusion feature map is obtained.  The structure of SPCModule  is shown in Figure \ref{fig8}.

\begin{figure}[ht]
\centering
\includegraphics [width=10cm]{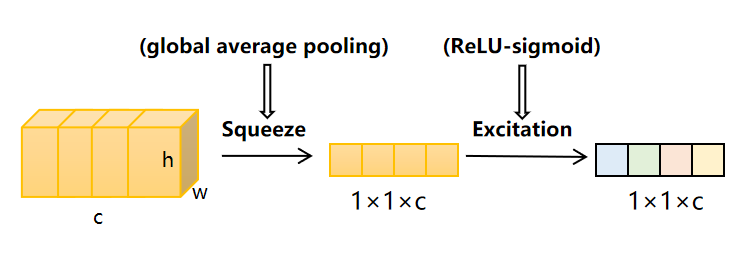}  
\caption{Structure of SEWeight module which includes the two most important components: Squeeze and Excitation.}
\label{fig9}
\end{figure}

On the other hand, the SEWeight module is designed to extract attention weights. Its structure, depicted in Figure \ref{fig9}, includes two essential components: Squeeze and Excitation. Global average pooling is applied to calculate the mean of each channel, compressing the feature maps into numerical information with a shape of $1\times 1 \times C$. This information is then processed through two linear layers with ReLU activation and a sigmoid activation function to obtain the weight for each channel.

\begin{figure}[ht]
\centering
\includegraphics [width=8cm]{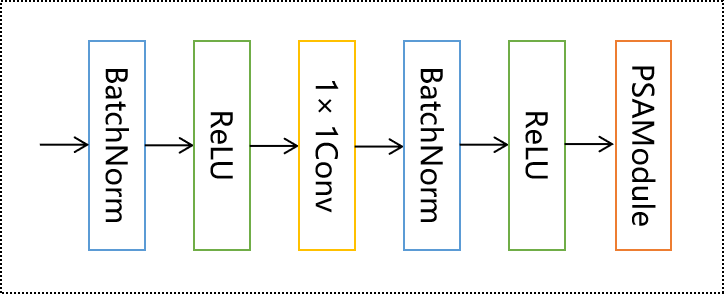}  
\caption{PSA-DenseNet model resulting from inserting the PSAModule into DenseNet121.}
\label{fig10}
\end{figure}

By combining the feature maps obtained after weighted channel attention, a new feature map containing multi-scale information is generated. In our implementation, we inserted the PSAModule into DenseNet121, replacing the $3\times 3$ convolution layer in the Dense Block. The resulting model, referred to as PSA-DenseNet, is depicted in Figure \ref{fig10}.

Overall, PSAModule introduces a powerful multi-scale approach based on channel attention. It significantly reduces the number of parameters and computational costs compared to other attention mechanisms. The SPCModule effectively captures spatial information at different scales, while the SEWeight module extracts channel attention. This integration enhances the model's ability to capture crucial features and establish long-term dependencies of information.

\subsubsection{MPSA-DenseNet model}
We are now ready to present the structure of our most powerful deep learning model: MPSA-DenseNet. MPSA-DenseNet integrates the PSAModule attention and multi-task learning methods, with the aim of exploring their combined effect on classification performance. The rationale behind this combination is that, in multi-task learning, there may be differences in the data and feature distributions across different tasks. By using PSAModule attention, the model can adaptively extract features of different scales and types based on the needs of each task, enhancing its representation ability.

The structure of MPSA-DenseNet is shown in Table \ref{table_1}. Its input is a sequence of MFCC feature maps, which are processed by a stack of dense blocks and transition layers. Each dense block contains several convolutional layers with batch normalization and ReLU activation functions. Meanwhile, the transition layers are used to reduce the spatial dimensions of the feature maps while increasing the number of channels.

In addition to the dense blocks and transition layers, MPSA-DenseNet also includes two auxiliary heads for the two tasks in our multi-task learning scenario. Each auxiliary head consists of a global average pooling layer followed by a fully connected layer with a softmax activation function. The outputs of the two heads are combined with a weight factor during training to achieve a balance between the two tasks.

In conclusion, our MPSA-DenseNet is a comprehensive model that leverages attention and multi-task learning techniques aiming at achieving superior classification performance on dataset. 

\begin{table}[ht]
  \centering
    \begin{tabular}{c|c|c}
      \toprule
      Layers & MPSA-DenseNet & Output Size \\
      \midrule
      Convolution & $7 \times 7$, stride 2, 64 channels & 32$\times $256 \\
      Block 1 & $[\begin{array}{c}
    1 \times 1 Conv \\
    PSAModule
    \end{array}] \times 6, 448 channels$ & 16$\times $128 \\       
      Transition Layer 1 & $[\begin{array}{c}
    1 \times 1 Conv \\
    AvgPool (size=2)
    \end{array}], 208 channels$ & 8$\times $64 \\
      Block 2 & $[\begin{array}{c}
    1 \times 1 Conv, 256 \\
    PSAModule
    \end{array}] \times 12, 976channels$ & 8$\times $64 \\
      Transition Layer 2 & $[\begin{array}{c}
    1 \times 1 Conv \\
    AvgPool (size=2)
    \end{array}] ,488 channels$ & 4$\times $32 \\
      Block 3 & $[\begin{array}{c}
    1 \times 1 Conv, 256 \\
    PSAModule
    \end{array}] \times 24, 2024 channels$ & 4$\times $32 \\
      Transition Layer 3 & $[\begin{array}{c}
    1 \times 1 Conv \\
    AvgPool (size=2)
    \end{array}] ,1012 channels$ & 2$\times $16 \\
      Block 4 & $[\begin{array}{c}
    1 \times 1 Conv, 256 \\
    PSAModule
    \end{array}] \times 16, 1396 channels$ & 2$\times $16 \\
    \hline
    \multirow{2}{*}{Classification Layer} & GlobalAvgPool(size=1) & 1$\times $1 \\
    \cline{2-3}
     &\multicolumn{2}{c}{6d/2d/5d fc, Softmax} \\
    \bottomrule
    \end{tabular}   
 \caption{Structure of MPSA-DenseNet model.\label{table_1}}
\end{table}  

\subsubsection{Evaluation Metrics}

We divide the dataset into three subsets: training, validation, and test sets, in the ratio of $6:2:2$. First, we train the models on the training set and then use their classification accuracy on validation set as the objective function to tune hyperparameter. The parameter setting that maximizes the accuracy on validation set is chosen. Utilizing these chosen parameter settings, we retrain the models on the union of training set and test set to increase the models' accuracy. The test set was used to evaluate the classification performance of the models. Despite attempts to balance the distribution of data labels during data collection, imbalances persisted, especially with age labels in multi-task learning methods where the label distribution was evident (see, Figure~\ref{figure1}). Therefore, relying solely on the accuracy of the model training is insufficient in expressing the classification performance of our model.

To ensure more reliable  comparison of performance of our model, we uses the $F_{\beta}$ score as an evaluation metric. $F_{\beta}$ score is an effective measure for imbalanced data sets in classification problems,  which is defined by

$$F_\beta = (1+\beta^2) \times \frac{\text{Precision} \times \text{Recall}}{(\beta^2 \times \text{Precision}) + \text{Recall}}.$$

The value of $\beta$ affects the proportion of prediction and recall in the evaluation index. 
To increase the proportion of prediction, the $F_{0.5}$ score with $\beta=0.5$ is chosen as the evaluation metric.

\subsubsection{Environment and Hyperparameters}

Our study uses the CUDA 11.0 environment, and MXNET framework which provides a variety of APIs and has the advantage of saving memory. Softmax Cross Entropy is used as the loss function of classification. To ensure a clear comparison of different models, each model is trained using the same set of hyperparameters. Specifically, the hyperparameters used in the models are 128 epochs, 16 batch sizes, and a learning rate of 0.0001. The optimization function used is Adam.

\section{Results}
In this section, we present the main results of our study, focusing on the performance of the three proposed deep learning models: Multi-DenseNet, PSA-DenseNet, and MPSA-DenseNet, compared to other models.

\begin{figure}[htbp]
\centering
\includegraphics[width=6cm]{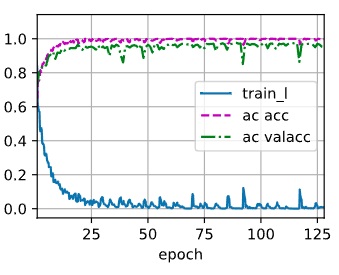}  
\includegraphics[width=6.7cm]{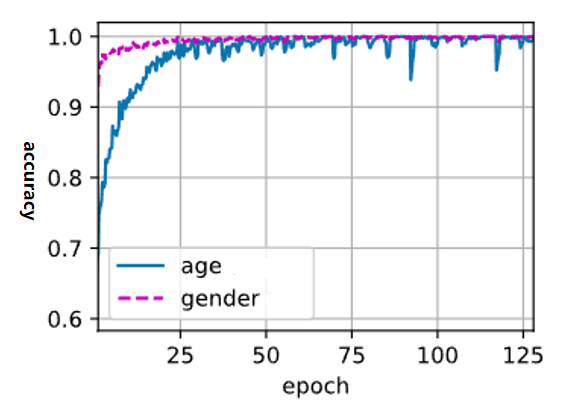}\\  
\includegraphics[width=6cm]{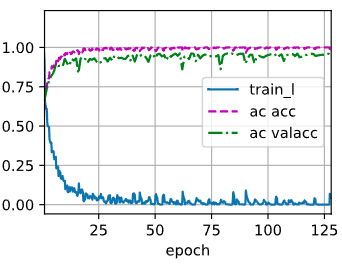}\\  
\includegraphics[width=6cm]{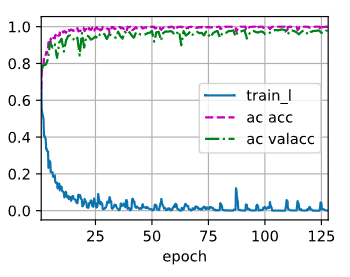}
\includegraphics[width=6.5cm]{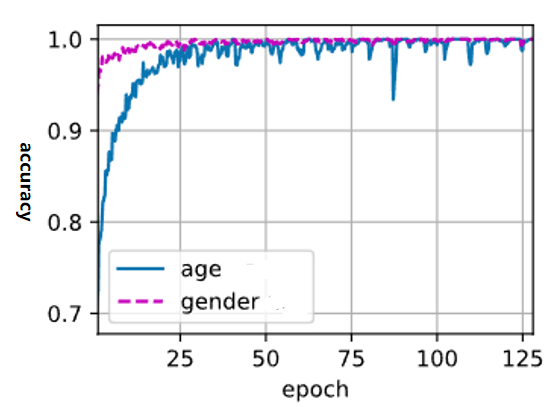}
\caption{The accuracy and loss function in identifying accents on training and validation sets (first column) and the accuracy of the gender and age auxiliary tasks on training set from Multi-DenseNet (first row), PSA-DenseNet (second row), and MPSA-DenseNet (third row).}
\label{fig15}
\end{figure}

Figure \ref{fig15} illustrates the results for Multi-DenseNet (first row), PSA-DenseNet (second row), and MPSA-DenseNet (third row). The sub-figures on the left in the first and third rows, as well as the sub-figure in the second row, depict the accuracy of accent identification on both the training and validation sets, plotted against the number of epochs. Additionally, the downward trend of losses on the training set is shown. On the right side, the sub-figures in the first and third rows display the training accuracy of the gender and age auxiliary tasks. It is worth noting that for the PSA-DenseNet model, since we did not incorporate multiple tasks, there is no figure depicting the training accuracy of the gender and age auxiliary task.

From the figure, we can observe that as the number of epochs increases, the accuracy of accent identification improves while the training loss decreases. This indicates that the models are learning and adapting to the accent classification task over time. Additionally, the training accuracy of the gender and age auxiliary tasks shows positive progress, demonstrating the effectiveness of the multi-task learning approach in these models.

\begin{figure}
\centering
\includegraphics[width=8cm]{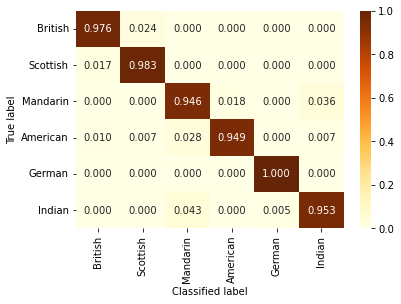}
\caption{Normalized confusion matrix of accent classification resulted from MPSA-DenseNet model on the test set}
\label{confusion_matrix}
\end{figure}

To better understand the quantitative results, let us examine the normalized confusion matrix of accent classification resulting from the MPSA-DenseNet model on the test set, as shown in Figure~\ref{confusion_matrix}. The diagonal elements represent the number of files correctly classified. While the accuracy for Mandarin and American accents is relatively lower compared to the other accents, it still exceeds $94\%$. Notably, the model achieved a remarkable $100\%$ accuracy in recognizing German accents.

Figure~\ref{confusion_matrix} also reveals that the most frequent error involves misclassifying Indian accents as Mandarin accents, occurring in $4.3\%$ of the cases. Other misclassifications are less than $3.6\%$. These results indicate that the MPSA-DenseNet model excels in recognizing accents, particularly German accents, and exhibits a low error rate for misclassification.

Although Multi-DenseNet and PSA-DenseNet did not perform as well as MPSA-DenseNet (as shown in Table \ref{table_4}), their performance remains competitive with other established models such as DenseNet 121 and EPSANet. Table \ref{table_4} displays the performance of all models on both the test and validation sets at the 128th epoch. It is evident that all indicators for Multi-DenseNet, PSA-DenseNet, and MPSA-DenseNet, except for the accuracy of PSA-DenseNet on the validation set, surpass those of DenseNet 121 and EPSANet. Notably, all indicators for MPSA-DenseNet are significantly higher than those of the other models.

\begin{table}[htbp]
  \centering
    \begin{tabular}{|l|c|c|c|c|}
        \toprule
        \diagbox{Model}{Evaluation} & Micro $F_{0.5}$ (Test set) & Macro $F_{0.5}$ (Test set) & Accuracy (Validation set) & Total Parameters\\
        \midrule
        DenseNet 121 & 93.6\% & 92.7\% & 93.9\% & 7.98M \\ 
        EPSANet & 94.3\% & 93.5\% & 94.3\% & 20.6M \\
        Multi-DenseNet & 95.3\% & 94.9\% & 95.1\% & 5.56M \\
        PSA-DenseNet & 94.8\% & 94.2\% & 93.8\% & 19.92M \\
        MPSA-DenseNet & 97.2\% & 96.4\% & 97.5\% & 19.64M \\
        \bottomrule
    \end{tabular}
    \hfill
 \caption{Performance comparison of each model on test and validation sets.\label{table_4}}
\end{table}

\section{Conclusion}

In this  study, we have introduced three novel models, namely Multi-DenseNet, PSA-DenseNet, and MPSA-DenseNet, designed to accurately classify English accents from both native and non-native speakers. By synergizing multi-task learning and the PSA module attention mechanism with DenseNet, we have achieved a a remarkable improvement in classification accuracy, especially with the MPSA-DenseNet model, which outperforms all other models.

Our findings underscore the power of combining different components, as each model performs better than using the components individually. The MPSA-DenseNet model shows exceptional generalization capabilities by achieving high accuracy across all six English accent types present in the dataset.

While we acknowledge the significant memory resources required for training the DenseNet architecture combined with multi-task learning and attention mechanism, we propose exploring lighter-weight network architectures, such as CondenseNet \cite{HLV}, in future research endeavors.

In summary, our study presents an efficient approach to English accent classification, paving the way for further research in this area. The results highlight the potential of MPSA-DenseNet as a promising solution for accent recognition tasks.


\begin{thebibliography}{00}


\bibitem {ICY}  Ittichaichareon, C., Suksri, S., $\&$ Yingthawornsuk, T.  Speech recognition using MFCC. {\em International conference on computer graphics, simulation and modeling} {\bf 9} (2012).
\bibitem {HMG}   Hossan, M. A., Memon, S., $\&$   Gregory, M. A. A novel approach for MFCC feature extraction. {\em 2010 4th International Conference on Signal Processing and Communication Systems.} 1-5 (2010).
\bibitem {AAR}    Aida-Zade,K.R., Ardil, C., $\&$ Rustamov, S.S. Investigation of combined use of MFCC and LPC features in speech recognition systems. {\em World Academy of Science, Engineering and Technology, 19,} 74-80 (2006).
\bibitem {PD}    Pedersen, C., $\&$ Diederich, J. Accent classification using support vector machines. {\em 6th IEEE/ACIS International Conference on Computer and Information Science (ICIS 2007)} 444-449 (2007).
\bibitem {RA}   Rizwan, M., $\&$ Anderson, D. V. A weighted accent classification using multiple words. {\em Neurocomputing, 277,} 120-128 (2018).
\bibitem {TG}     Tang, H., $\&$ Ghorbani, A. A. Accent classification using support vector machine and hidden markov model. {\em Advances in Artificial Intelligence: 16th Conference of the Canadian Society for Computational Studies of Intelligence, AI 2003, Halifax, Canada, June 11–13, 2003, Proceedings 16.} 629-631. Springer Berlin Heidelberg (2003).
\bibitem {HLZ}    Hou, J., Liu, Y., Zheng, T. F., Olsen, J., $\&$ Tian, J. Multi-layered features with SVM for Chinese accent identification. {\em 2010 International Conference on Audio, Language and Image Processing.} 25-30 (2010).
\bibitem {AH}   Angkititrakul, P., $\&$ Hansen, J. H. Advances in phone-based modeling for automatic accent classification. {\em IEEE transactions on audio, speech, and language processing, 14(2),} 634-646 (2006).
\bibitem {KK}   Kumpf, K., $\&$ King, R. W. Automatic accent classification of foreign accented Australian English speech. {\em Proceeding of Fourth International Conference on Spoken Language Processing. ICSLP'96} (Vol. 3, pp. 1740-1743). IEEE (1996).
\bibitem {BWE}     Bird, J. J., Wanner, E., Ekárt, A., $\&$ Faria, D. R. Accent classification in human speech biometrics for native and non-native english speakers. {\em Proceedings of the 12th ACM International Conference on PErvasive Technologies Related to Assistive Environments} 554-560 (2019).
\bibitem {YMV}     Jiao, Y., Tu, M., Berisha, V., $\&$ Liss, J. M. Accent Identification by Combining Deep Neural Networks and Recurrent Neural Networks Trained on Long and Short Term Features. {\em Interspeech} 2388-2392 (2016).
\bibitem {KMY}  Chionh, K., Song, M., $\&$ Yin, Y. Application of Convolutional Neural Networks in Accent Identification. {\em Project Report, Carnegie Mellon University, Pittsburgh, Pennsylvania} (2018).
\bibitem {ZTA}     Al-Jumaili, Z., Bassiouny, T., Alanezi, A., Khan, W., Al-Jumeily, D., $\&$ Hussain, A. J. Classification of Spoken English Accents Using Deep Learning and Speech Analysis. {\em Intelligent Computing Methodologies: 18th International Conference, ICIC 2022, Xi'an, China, August 7–11, 2022, Proceedings, Part III} 277-287. Cham: Springer International Publishing (2022).
\bibitem {QHY}     Gao, Q., Wu, H., Sun, Y., $\&$ Duan, Y. An end-to-end speech accent recognition method based on hybrid ctc/attention transformer asr. {\em ICASSP 2021-2021 IEEE International Conference on Acoustics, Speech and Signal Processing (ICASSP)} 7253-7257. IEEE (2021).
\bibitem {YHD}     Zeng, Y., Mao, H., Peng, D., $\&$ Yi, Z. Spectrogram based multi-task audio classification. {\em Multimedia Tools and Applications, 78,} 3705-3722 (2019).
\bibitem {SAA}  Weinberger, Steven. Speech Accent Archive. George Mason University. Retrieved from http://accent.gmu.edu (2015).
\bibitem {VCTK}     Yamagishi, Junichi; Veaux, Christophe; MacDonald, Kirsten. CSTR VCTK Corpus: English Multi-speaker Corpus for CSTR Voice Cloning Toolkit (version 0.92), [sound]. University of Edinburgh. The Centre for Speech Technology Research (CSTR). https://doi.org/10.7488/ds/2645 (2019).
\bibitem {MRL}     McFee, B., Raffel, C., Liang, D., Ellis, D. P., McVicar, M., Battenberg, E., $\&$ Nieto, O. librosa: Audio and music signal analysis in python. {\em Proceedings of the 14th python in science conference.} Vol. 8, pp. 18-25 (2015).
\bibitem {TS}  Tomar, Suramya. Converting video formats with FFmpeg. {\em Linux journal, 2006} (146),10 (2006).
\bibitem {BE}     Brigham, E. O. {\em The fast Fourier transform and its applications.} Prentice-Hall, Inc. (1998).
\bibitem {GZL}  Huang, G., Liu, Z., Van Der Maaten, L., $\&$ Weinberger, K. Q. Densely connected convolutional networks. {\em Proceedings of the IEEE conference on computer vision and pattern recognition} 4700-4708 (2017).
\bibitem {CR}   Caruana, R. Multitask learning. {\em Machine learning, 28,} 41-75 (1997).
\bibitem {CM}  Crawshaw, M. Multi-task learning with deep neural networks: A survey. {\em arXiv preprint arXiv:2009.09796} (2020).
\bibitem {HKJ}  Zhang, H., Zu, K., Lu, J., Zou, Y., $\&$ Meng, D. EPSANet: An efficient pyramid squeeze attention block on convolutional neural network. {\em Proceedings of the Asian Conference on Computer Vision} 1161-1177 (2022).
\bibitem {HLV}     Huang, G., Liu, S., Van der Maaten, L., $\&$ Weinberger, K. Q. Condensenet: An efficient densenet using learned group convolutions. {\em Proceedings of the IEEE conference on computer vision and pattern recognition} 2752-2761 (2018).


\end{thebibliography}
\end{document}